# Retraction mechanism of soft torus robot with a hydrostatic skeleton

Tomoya takahashi, Masahiro Watanabe, Kenjiro Tadakuma, Masashi Konyo and Satoshi Tadokoro,
*Tohoku University*

*Abstract*—Soft robots have attracted much attention in recent years owing to their high adaptability. Long articulated soft robots enable diverse operations, and tip-extending robots that navigate their environment through growth are highly effective in robotic search applications. Because the robot membrane extends from the tip, these robots can lengthen without friction from the environment. However, the flexibility of the membrane inhibits tip retraction. Two methods have been proposed to resolve this issue; increasing the pressure of the internal fluid to reinforce rigidity, and mounting an actuator at the tip. The disadvantage of the former is that the increase is limited by the membrane pressure resistance, while the second method adds to the robot complexity. In this paper, we present a tip-retraction mechanism without bending motion that takes advantage of the friction from the external environment. Water is used as the internal fluid to increase ground pressure with the environment. We explore the failure pattern of the retraction motion and propose plausible solutions by using hydrostatic skeleton robot. Additionally, we develop a prototype robot that successfully retracts by using the proposed methodology. Our solution can contribute to the advancement of mechanical design in the soft robotics field with applications to soft snakes and manipulators.

*Keywords—Soft Robot Materials and Design, Mechanism Design*

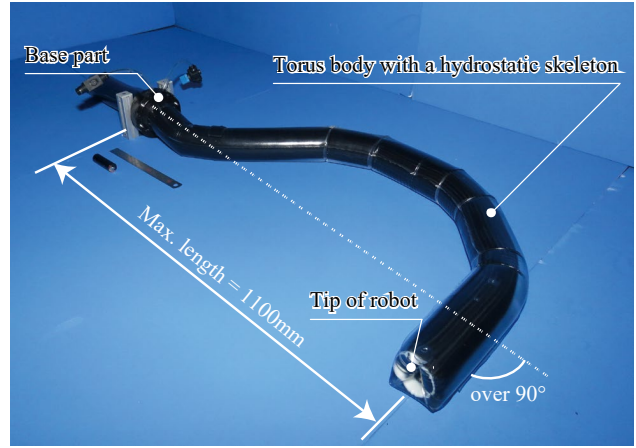

Fig. 1 The torus-type robot we proposed can retract its body from the tip even if it curves smaller over 90°.

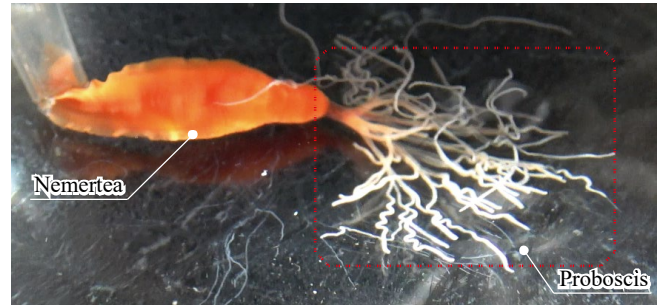

Fig.2 Nemertea has a tip-extending mechanism in their proboscis that can branch, extend, and bend. They can also retract their proboscis by pulling it from the tip.

## I. INTRODUCTION

In recent years, soft robots have received increasing attention. Particularly, long multi-segmented robots are highly articulated to enable diverse operations (e.g., varied task, wide-range movement) [1] and applications (e.g., robot arms and snake-type robots) [2][3][4]. Additionally, articulated robots can deform the robot to a smaller radius of curvature by increasing the number of joints, shortening the length of each joint, and increasing the overall length. For example, in a search robot navigating in a narrow area, the smaller the curvature is, the more the movement range increases [6] [7]. If a long segmented robot is composed of a soft body, it can be regarded as a highly articulated and extremely small link, and can be deformed with a smaller curvature [5]–[7].

Among long soft robots, tip-extending robot have been proposed [8]–[15] and have shown high effectiveness, especially in the field of search robotics. Tip-extending robots carry a mechanism that allows the membrane to grow from the end point. The advantage of this mechanism is that the tip can extend without friction from sliding against the environment.

Most of the tip-extending search robots have a single snake structure without branches in the main body. If branches are attached at the end of the extension, it is possible to search exhaustively and efficiently using a single actuator. As a latest. research, we have composed the branched tip extension mechanism [18]. Lucarotti et al. also proposed the development of a robot that extends and branches out, inspired from plant roots. However, these researches could not successfully retract the tip yet.

Creatures, such as some type of nemertea, have been reported to carry a tip-extending mechanism in their proboscis that can branch, extend, bend, and retract (Fig.2) [19]–[21]. They use this structure for predation and locomotion. Nemertea can extend and retract their proboscis repeatedly by applying internal pressure [22][23]. These features are not achieved by conventional tip-extending robots. The most challenging part is the retraction mechanism because tip-

Tomoya Takahashi, Masahiro Watanabe, Kenjiro Tadakuma, Masashi Konyo, and Satoshi Tadokoro are with the Graduate school of Information Sciences, Tohoku University, Japan (Corresponding author: Kenjiro Tadakuma (email: tadakuma@rm.is.tohoku.ac.jp).).

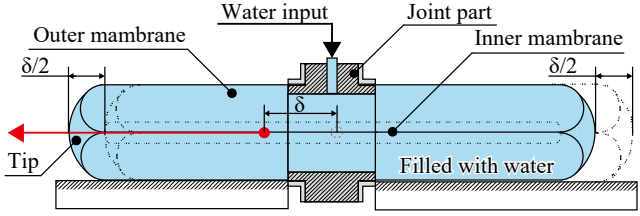

Fig.3 Schematic of a torus tip-extending robot

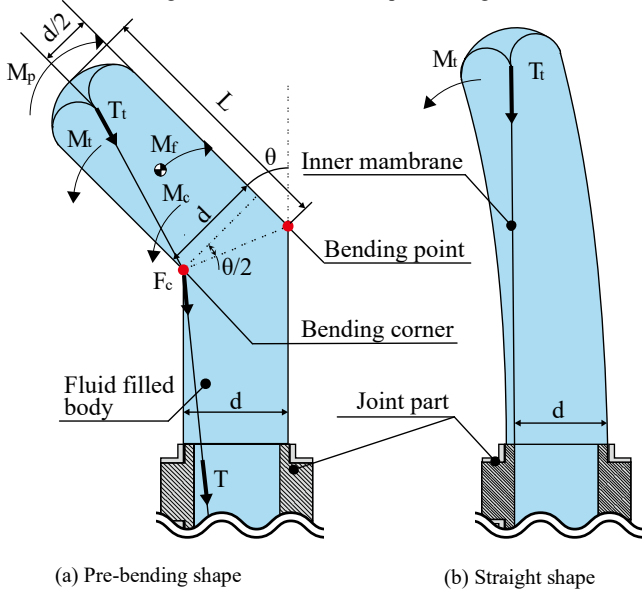

(a) Pre-bending shape  (b) Straight shape

Fig.4 Free body diagram with the forces acting on the torus body

extending robots are often constructed with a soft membrane that allows the extension and folding of the tip. When the membrane is pulled during retraction, buckling occurs at the base where large moment is applied [25].

The goal of this study is to realize a tip-extending mechanism that can branch, extend, curve, and retract inspired by the proboscis of nemertea. Further, we analyze the deformation modes when retracting the tip of a simple straight tip-extending robot.

The main contributions of this study are the exploration of the retraction fails modes of a torus-type tip-extending robot and the conditions for its occurrence and the development of a method that prevents buckling even when the robot body is composed of a soft membrane. This paper is structured as follows. In Section II, we discuss the factors that cause buckling and study the mechanical conditions the retraction. In Section III, we present the development of a prototype design. In Section IV, we perform experiments to validate the effectiveness of the proposed mechanism, and in Section V includes our conclusions and scope of future work.

## II. PRINCIPLE

Alison et al. proposed two methods for successfully retracting the tip. First, they increased the stiffness of the robot by applying high pressure [24]. However, retraction was only possible for high pressure resistance which was difficult to achieve owing to the thin robot material. Second, an actuator was mounted at the tip to fold back the membrane of the robot. Nevertheless [25] [26], this solution complicates the structure of robot, especially if it is branched. In this study, we consider the retraction mechanism of the robot at low pressure to enable the retraction without depending on the pressure of the working fluid.

### A. Prototype and model

Fig. 3 shows the structure of a torus-type tip-extending robot. Both ends of the robot are folded back from the outer to the inner membrane. The inner membrane is connected on the inside, while both ends of outer membrane are fixed to the center joint. When the inner membrane is pulled by a factor $\delta$ to the left, the length of the right side contracts by $\delta/2$, and the left side extends by $\delta/2$, as illustrated in Fig. 3. Then, a joint is provided in the middle to connect with the outer membrane, and the inside is filled with fluid injected from the joint.

By symmetrically connecting the two tip-extending structures, we achieve certain advantages compared to the single mounting structure. First, since there is no volume change while the robot moves, there is no pressure change when retracting. Second, the force required for contraction is small and constant because tension is applied to both ends of the inner membrane and is, thus, balanced.

Conventionally, robots are pneumatically driven, but in this paper, we use water as the internal fluid. Therefore, the lower side of the outer membrane is always in contact with the ground surface owing to the added weight. This is discussed in detail in Section II-C (iii).

### B. Forces acting on the torus body

The straight torus-type robot presented in the previous section can be bent in the middle, while keeping its shape owing to friction from the ground because the pressure of the working fluid is low and ground pressure is large. Fig. 4 shows the free body diagram when elbow of the robot is retracted from the tip. The moments generated in the body are discussed as follows.

Note that the model assumes a circular cross section, and the inner membrane can be regarded as a wire without volume. It is also assumed that the membrane does not expand or contract owing to the generated force.

(i.) $M_t$ : Bending moment owing to tip tension $T$

The moment when tension $T_t$ is generated at the tip as shown in Fig. 4 and Fig.6 (a) is described as

$$M_t = r_t \cdot T_t = dLT_t / \sqrt{\left(\frac{d}{2}\right)^2 + L^2} \quad (1)$$

where d is the diameter of the body. Then, using the force $F_0$ required to fold the tip back, $M_t$ can be expressed as

$$M_t = dLF_0 / \sqrt{\left(\frac{d}{2}\right)^2 + L^2} \qquad (2)$$

*(ii.) $M_b$: Bending moment owing to friction generated at the bending corner*

When the inner membrane is bent along the bending corner, the bending moment is generated by the frictional force. According to Euler's belt theory, when the inner membrane bent at a pre-bending angle $\theta$ starts to slip, the tension $T_b$ at the base is

$$T_b = F_0 e^{\mu_b \theta}, \qquad (3)$$

Where $\mu_b$ is the coefficient of friction between membranes. The force $F_c$ applied from the inner to the outer membrane can be expressed as follows

$$F_c = T_b - T_t = F_0(e^{\mu_b \theta} - 1). \qquad (4)$$

If the base side of the robot is sufficiently long, $F_f$ can be working in parallel with the outer membrane at the root, so the bending moment $M_c$ owing to friction is given as follows

$$M_c = F_0 d(e^{\mu_b \theta} - 1)/cos(\theta/2). \qquad (5)$$

*(iii.) $M_f$: Moment of resistance generated by friction with the ground*

The maximum value of the bending moment owing to friction is expressed by the following equation.

$$M_{fmax} = \frac{\rho \mu g l^2}{2}, \qquad (6)$$

Where $\rho, \mu, g$ are the mass of the body per unit length, the coefficient of friction between the body and the ground, and the gravitational acceleration, respectively.

*(iv.) $M_p$: Moment to return to a straight line by pressure*

We experimentally calculated the moment $M_p$, at which the body tries to return to the original state by applying pressure. Note that it is small because the pressure is low. A detailed explanation of this is provided in II-C (iii)

*C. Modes of deformation*

It was found that the mode of retraction failure was not uniform but changed to various shapes just by pulling the inner membrane when the torus-type tip-extending robot was operated with water. Through experiments, we found that the deformation pattern of the robot was determined by its initial shape. The classification of this deformation pattern and its cause are summarized below.

*(i.) Straight-bending*

In this phenomenon, when tension is applied to the inner membrane of a body that is linear at the initial state, buckling occurs at a distance $L$ from the tip.

When the body is straight, there is no bending moment. However, when the body is curved at a constant curvature, as shown in Fig. 4(b), moment is generated in the direction in which the body is curved and the bent. The moment $T_t$ from

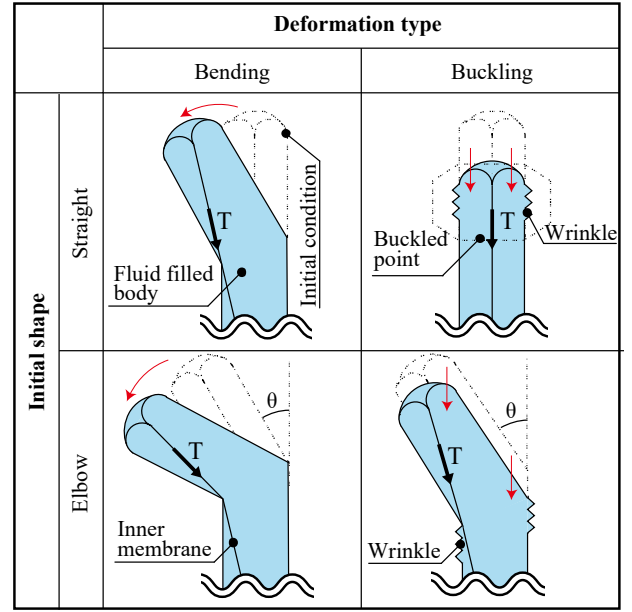

Fig.5 Modes of torus robot deformation

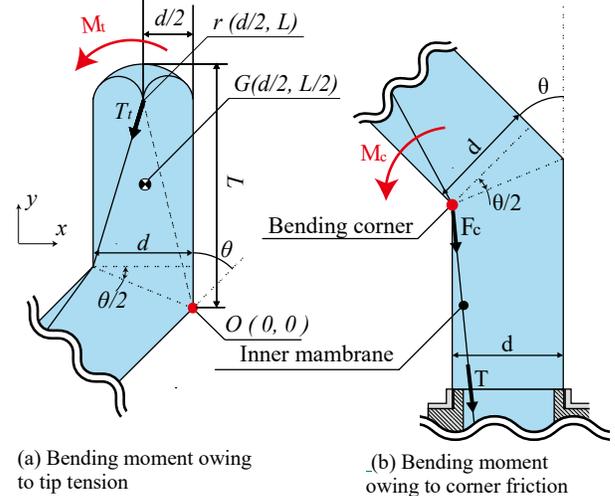

(a) Bending moment owing to tip tension    (b) Bending moment owing to corner friction

Fig.6 Schematic of bending moment

the tension applied to the tip of the robot generated at this time is expressed as

$$M_t = T_t d. \qquad (7)$$

Moreover, if the distance from the bend corner to the tip is $L$, buckling occurs at the base when $M_t > M_f$, so the equation of $T_t$ at which bend occurs is expressed as

$$T_t > \frac{\rho \mu g L^2}{2d}. \qquad (8)$$

This mode was not confirmed when using water as the internal fluid and operating at low internal pressure and high ground pressure, but it was observed when air was injected at high pressure.

*(ii.) Straight-buckling*

When tension is applied to the inner membrane, bellow-like deformation occurs before the tip folds back. This phenomenon occurs at low pressure, contrary to (i). When the inner membrane is pulled from the tip and the body is deformed in a bellows shape, the volume of the body decreases the force $F_b$ required to buckle the body from the tip. Therefore, the condition under which straight-buckling occurs is

$$T_t > F_0 > F_b. \qquad (9)$$

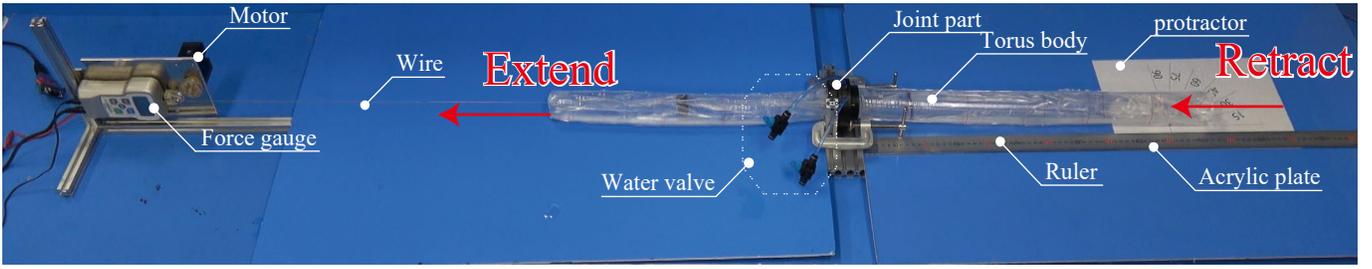

Fig. 7  Prototype of torus-type robot

*(iii.) Elbow-bending*

This phenomenon is a phenomenon in which $\theta$ becomes larger when a force is applied to the inner membrane of the body that is originally bent at an angle $\theta$. Once bend occurs, $\theta$ increases and the retraction fails. The condition under which bending occurs at a pre-bending angle $\theta$ is determined by the distance $L$ from the bending point. The derivation of this condition is as follows. First, when the body is bent, the forces applied to it are as shown in Fig. 3. At this time, the condition that the tip of the body slips can be expressed by the balance of the moments:

$$M_t + M_c > M_{fmax} + M_p. \qquad (10)$$

Combining Eqs. (2), (5), and (6), we find the condition for elbow-bending:

$$dLF_0 / \sqrt{\left(\frac{d}{2}\right)^2 + L^2} + F_0 d(e^{\mu_b \theta} - 1) > \frac{\rho \mu g L^2}{2} + M_p \qquad (11)$$

To prevent elbow-bending, it is conceivable to increase $M_f$ or decrease $M_t, M_c$. In this study, this was done by using water as the internal fluid because the density $\rho$ is increased when using water compared to using air. Additionally, the incompressible nature of water, enables filling the interior with liquid at a low pressure, so that the tensions $T_t$ reduced and $M_t$ becomes smaller than $M_f$.

*(iv.) Elbow-buckling*

When the body bends at large angles, the force applied to the bending corner and base increases, and the body is buckled at the base before bending or retracting. Simultaneously, the tip of the body rotates in the direction in which $\theta$ decreases. When the force $F_c$ generated at the bending corner is applied and the tip of the body of length $L$ slips, the following relationship is established:

$$F_c L > M_{fmax} = \frac{\rho \mu g L^2}{2} \qquad (12)$$

$$F_c > \frac{\rho \mu g l}{2}. \qquad (13)$$

The condition for generating elbow-buckling is expressed by the following equation:

TABLE II. SUCCESS RATE OF RETRACTION THROUGH THE BENDING POINT

| Length | Diameter | Weight | Wire diameter |
|---|---|---|---|
| 1100 mm | 50 mm | 2318 g (at 100%) | 0.1 mm |

$$F_c > \frac{\rho \mu g l}{2} + F_b. \qquad (14)$$

Here, $F_c$ is the frictional force generated at the bent part, and thus can be expressed by the following equation:

$$F_c = F_0(e^{\mu_b \theta} - 1). \qquad (15)$$

Therefore, for elbow-buckling to occur, the bending angle $\theta$ and the bending tip length $L$ must satisfy the following formula:

$$F_0(e^{\mu_b \theta} - 1) > \frac{\rho \mu g l}{2} + F_b. \qquad (16)$$

*D. Retract through the bending point*

So far, we have described the retraction method from the tip to the bending point. In this section, we focus on the method of retraction through the bending point. The elbow-buckling discussed in the previous section can be avoided by reducing the pre-bending angle. However, the condition we calculated indicates that the elbow-bend always occurs during the retraction process. The reason is that independent of whether the pre-bending angle diminishes or whether the retraction is successful, $L$ approaches 0. Consequently, $M_f$ becomes smaller than the bending moment. To avoid this phenomenon, it is necessary to increase the pressure and corresponding moment $M_p$. However, there is a limit to the increase in pressure as explained in section I. Therefore, in this study, the strength when $L$ is small was reinforced by a mesh tube to successfully retract the tip through the bending part. The method is explained in detail in section III-B.

## III. DESIGN

*A. Torus-type tip-extending robot*

A first prototype of the proposed torus-type robot shown in Fig. 7 was made for performing our experiments. A 0.1 mm polyurethane sheet was used as a flexible material, and two sheets were welded and formed into a tube. This polyurethane tube was folded back to form a torus structure. The ends of the outer membrane were connected with a joint made by a 3D printer. Then, water was injected into the tube from the joint. The dimensions of the body of the robot are shown in Table I.

*B. Torus-type robot with guide tube*

To make the torus-type robot less likely to buckle, we implemented a guide tube with a low coefficient of friction

(Fig. 1). This is the same configuration as that of the prototype with the guide tube inserted between the inner and outer membranes. The guide tube does not fold back with the membrane at the tip but slides along the direction of extension and contraction of the tip.

This structure has two advantages compared to the prototype. First, a reduction in friction between the membranes is achieved. As described in section II-A, it is important to reduce the friction at the bending part for reducing the bending moment. Second, the stiffness of the robot is increased by adding the tube. The guide tube is made out of a high stiffness material, stiffer than that of the membrane, because it does not need to be folded back and must keep its shape while the robot moving. Additionally, to reinforce the body to prevent deformation, we added a nylon mesh tube. Nylon was selected because it is slippery and can bend greatly while maintaining radial strength. The mesh tube was inserted while compressing it in the longitudinal direction by half the length. It can be extended to the length of the mesh tube to be the same as the robot owing to its high elasticity. Furthermore, an ABS ring made by a 3D printer was fixed to both ends of the mesh tube to prevent it from being folded.

## IV. EXPERIMENTAL PROCEDURE

### A. Experimental setup

The experimental setup is illustrated in Fig. 8. A wire is welded to the center of the inner membrane of the torus-type robot and is connected to the motor (Dynamixel XH540-V150-R, South Korea) via a pulley attached to the force gauge (FGP-5, Nidec-Shimpo Corp., Japan). When the motor was rotated and the wire was wound twice, wire tension was applied to the force gauge. We logged this tension until the motor stopped.

In this experiment, we used volume filling rate to measure the filling amount of water inside the robot. The volume filling rate is defined as the ratio of the water volume to the maximum volume, which is estimated from the model described in III-A as 100%. This volume is measured by the mass of the water inside the robot. Note that the volume filling rate can be more than 100% because the film can stretch.

The most important part of the failure retraction modes is to prevent elbow-bending. Straight-bending and straight-buckling do not occur if the volume filling rate of water is sufficient, and elbow-buckling can be avoided by decreasing the bending angle. Therefore, in what follows, we experimentally verify the conditions for the generation of elbow-bending.

### B. Results

*(i.) Measurement of the buckling force*

To confirm the conditions that cause buckling, we conducted the experiment shown in Fig. 8 (b). We installed the torus-type robot and measured the maximum force when pushing 10 mm from the tip by using a force gauge attached on the guide rail. We measured the force by using four different volume filling ratios, i.e., 95%, 100%, 110%, and 120%. The average value of 10 trials was defined as the buckling force $F_b$. The lowest pressure corresponds to the minimum value for retraction.

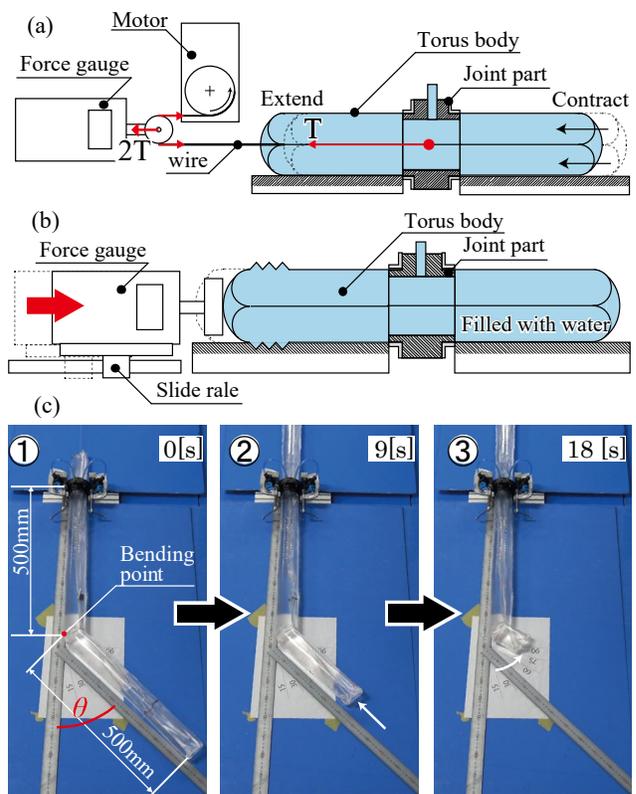

Fig.8(a) Experimental set up; (b) Measurement of buckling force; (c) Measurement of tension at elbow shape

*(ii.) Tension owing to the pre-bending angle*

To verify the tension effect on the body as shown in II-B, we measured the change in tension caused by the change in the pre-bending angle. We setup the torus-type robot as shown in Fig. 8 (a). Then, we fixed the joint to the floor and let one side of the body rest on the acrylic plate. The body kept its elbow shape because of the friction from the acrylic plate. In this experiment, the initial shape was such that the body was bent at 500 mm from the joint and the length from the bending point to the tip was 500 mm. When the motor was rotated at a constant speed (approx. 44 mm/s), the wire connected to robot and motor pulled inner membrane to folded buck the tip. It stopped when the body bends. The tension was measured with a force gauge, and the average value of the tension for 10 seconds after applying the tension was defined as the tension $T_m$ applied to the inner membrane. The balance of the force applied to the inner membrane is as follows.

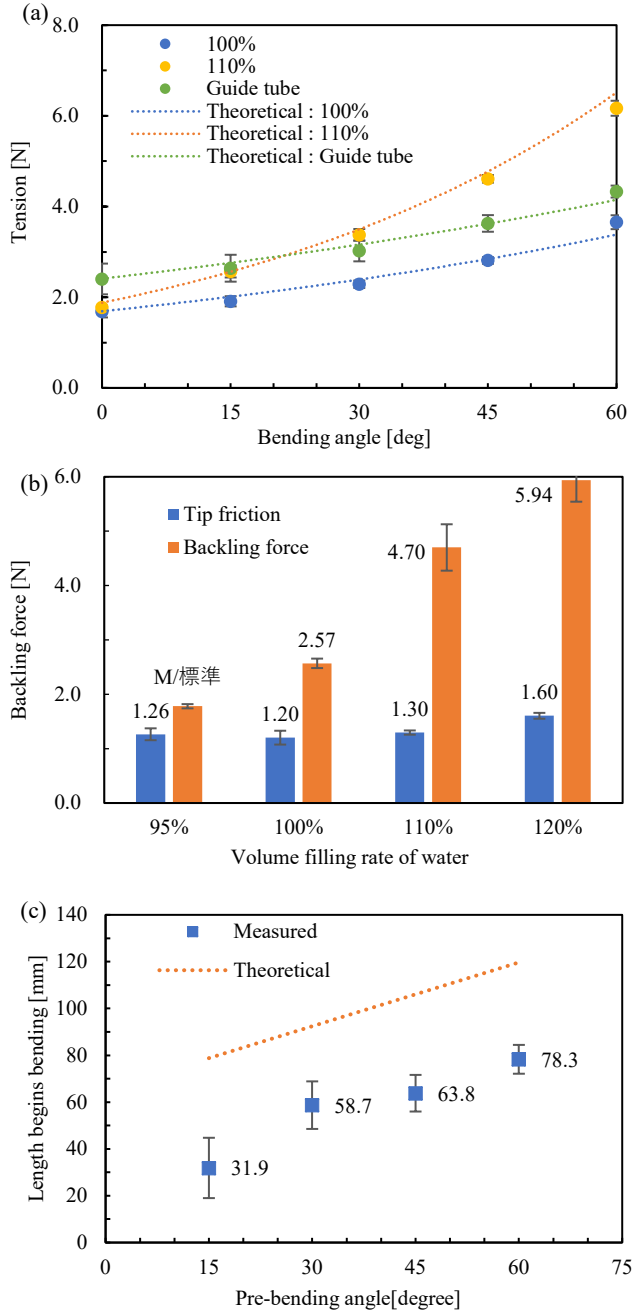

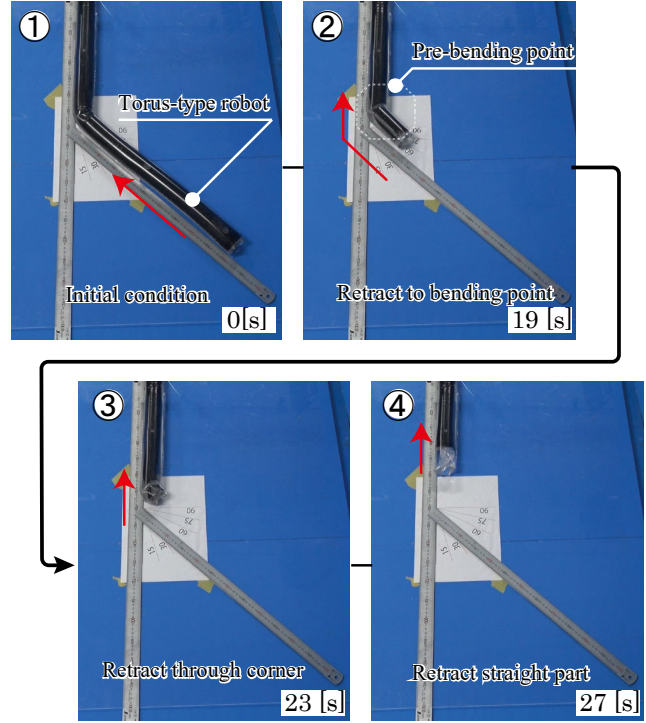

Fig. 10 Successful retraction at a pre-bending angle $\theta = 45°$

Fig. 9 (a) Comparison of the retraction resistance caused by the pre-bending angle; (b) Comparison between the folding resistance and buckling force; (c) Buckling start length.

$$T_m = F_0 e^{\mu_b \theta}. \tag{18}$$

We determined the coefficient of friction between the films at the bending corner by using the measured value of $T$ and Eq. (18). The measured tension increased exponentially with the angle, as shown in Fig. 9 (a). The coefficient of friction was determined as 0.55 to be in line with the measured values. In the case the mesh tube was inserted, the folding back resistance (the force for $\theta = 0$) increased owing to the restoring force of the tube. However, the coefficient of friction was 0.45, indicating that the tube was more slippery than the urethane film.

In this experiment, when the pre-bending angle was 15° to 45°, the tip of the robot was successfully retracted more than 400 mm after that bend occurred. At 60°, buckling occurred first after the tip retracted a few millimeters. Then, bending occurred after further retraction. At 75°, buckling occurred without retraction.

*(iii.) Measurement of folding resistance*

To verify the straight-buckling conditions explained in II-C, we measured the folding resistance caused by the difference in the water filling rate. Following the same experimental procedure as in (ii), we measured the tension at a pre-bending angle of 0°. The measured tension $T_m$ is the same as the folding resistance $F_0$.

Fig. 9 (b) shows the folding resistance when the internal pressure is changed. The folding resistance is smaller than the experimentally determined buckling force between the 95% and 120% volume filling. Then, buckling did not occur under these conditions.

*(iv.) Bend start length measurement at elbow-bending*

The elbow-bending condition explained in II-C was experimentally verified. The length $L$ from the bending point to the tip when bending occurred was measured at the same time as the experiment described in (ii). $L$ was determined the average of 10 measurements.

Fig. 9 (c) shows a comparison between the measured and theoretical value of the bending start length obtained from Eq. (12). The coefficient of friction between the acrylic plate and the polyurethane was 0.81, which was experimentally observed. The membrane was assumed to be sufficiently lighter than the internal fluid. The bending start length was actually much smaller than the theoretical value.

*(v.)* *Success rate of retraction through the bending point*

To verify the effectiveness of the guide tube for retraction through the bending point described in II-D, we compare the success rate for retracting (Table II). The success rate was calculated as the number of successful retractions through the bending point among 10 trials, as described in (iv). The results in Table II indicate that the success rate increases when the guide tube is inserted. Fig. 10 shows the return operation (also show in supplementary video).

TABLE II. SUCCESS RATE OF RETRACTION THROUGH THE BENDING POINT

| Only water | 15° | 30° | 45° | 60° |
|---|---|---|---|---|
| 100% | 0 | 0 | 0 | 0 |
| 110% | 100 | 0 | 0 | 0 |
| With guide tube | 100 | 100 | 30 | 0 |

## V. DISCUSSION

Through this study, we present practical findings regarding a design method that aims to improve the success rate of retraction in a tip-extending robot. We suggest the following methods for improving the success rate for tip retraction.

・Lower the folding resistance at the tip to prevent buckling at the bending point. For example, use a thinner membrane for the robot body or use a softer membrane with a lower coefficient of friction.

・Reduce the friction at the bending point to limit buckling and bending by changing to a more stretchable membrane material or to a guide tube material with a low coefficient of friction.

・Increase the density of the inner fluid to increase the ground pressure between the robot and the floor to prevent slippage.

・Reinforce the guide tube to support body stiffness by choosing a stiff and slippery material and further compressing the guide mesh tube.

## ACKNOWLEDGMENT

Advice and comments given by prof. Kajihara and his laboratory students (Hokkaido University) has been a great help in observation of nemertea.

## VI. CONCLUSION

Summing up, in this study, we investigated the retraction failure through theoretical calculations and experimental procedures based on a torus-body tip-extending robot design. Additionally, we derived the conditions for the occurrence of buckling and bending. It was found that the retraction failed when the bending moment from internal tension exceeded the maximum resistance moment from the friction between the membrane and the environment. The length at which the bend started was about 40 mm shorter than the theoretical value. This is thought to be due to the increase in moment due to pressure and the rigidity of the membrane that were not considered.

Further, we proposed three methods for successfully retraction: (1) Increase the resistance moment from friction with the external environment, (2) minimize the internal frictional resistance as much as possible, and (3) reinforce the robot body by inserting a guide tube. Certainly, inserting a tube enables retraction through the bending point. Thereby, retraction in a curved shape, as shown in Fig. 1, was also successful.

The phenomena explored in this study, such as buckling and bending, do not necessarily mean retraction failure but show that one degree of freedom can allow a variety of actions depending on the conditions. Namely, it can also be used as a function that can generate a large force to the bending point. If this function is applied, not only a search robot, but also applications such as grippers and deployment mechanisms can be considered. In future work, we intend to establish the design policy of the guide tube and develop a branch structure.

## REFERENCES

[1] M. Mahvash and M. Zenati, "Toward a hybrid snake robot for single-port surgery," in 2011 Annual International Conference of the IEEE Engineering in Medicine and Biology Society, Boston, MA, 2011, pp. 5372–5375, doi: 10.1109/IEMBS.2011.6091329.

[2] S. Hirose and H. Yamada, "Snake-like robots [Tutorial]," IEEE Robot. Automat. Mag., vol. 16, no. 1, pp. 88–98, Mar. 2009, doi: 10.1109/MRA.2009.932130.

[3] J. C. McKenna et al., "Toroidal skin drive for snake robot locomotion," in 2008 IEEE International Conference on Robotics and Automation, Pasadena, CA, USA, 2008, pp. 1150–1155, doi: 10.1109/ROBOT.2008.4543359.

[4] A. Horigome, H. Yamada, G. Endo, S. Sen, S. Hirose, and E. F. Fukushima, "Development of a coupled tendon-driven 3D multi-joint manipulator," in 2014 IEEE International Conference on Robotics and Automation (ICRA), Hong Kong, China, 2014, pp. 5915–5920, doi: 10.1109/ICRA.2014.6907730.

[5] M. Takeichi, K. Suzumori, G. Endo, and H. Nabae, "Development of Giacometti Arm With Balloon Body," IEEE Robot. Autom. Lett., vol. 2, no. 2, pp. 951–957, Apr. 2017, doi: 10.1109/LRA.2017.2655111.

[6] S. Tadokoro et al., "Application of Active Scope Camera to forensic investigation of construction accident," in 2009 IEEE Workshop on Advanced Robotics and its Social Impacts, Tokyo, Japan, 2009, pp. 47–50, doi: 10.1109/ARSO.2009.5587076.

[7] K. Hatazaki, M. Konyo, K. Isaki, S. Tadokoro, F. Takemura, "Active scope camera for urban search and rescue", 2007 IEEE/RSJ International Conference on Intelligent Robots and Systems, San Diego, CA, USA, 2007, pp. 2596–2602, doi: 10.1109/IROS.2007.4399386.

[8] H. Tsukagoshi, N. Arai, I. Kiryu, and A. Kitagawa, "Smooth creeping actuator by tip growth movement aiming for search and rescue operation," in 2011 IEEE International Conference on Robotics and Automation, Shanghai, China, 2011, pp. 1720–1725, doi: 10.1109/ICRA.2011.5980564.

[9] H. Tsukagoshi, A. Kitagawa, and M. Segawa, "Active Hose: an artificial elephant's nose with maneuverability for rescue operation," in Proceedings 2001 ICRA. IEEE International Conference on Robotics and Automation (Cat. No.01CH37164), Seoul, South Korea, 2001, vol. 3, pp. 2454–2459, doi: 10.1109/ROBOT.2001.932991.

[10] I. Kiryu, H. Tsukagoshi, and A. Kitagawa, "GROW-HOSE-I: A HOSE TYPE RESCUE ROBOT PASSING SMOOTHLY THROUGH NARROW RUBBLE SPACES," Proceedings of the JFPS International Symposium on Fluid Power, vol. 2008, no. 7–3, pp. 815–820, 2008, doi: 10.5739/isfp.2008.815.

[11] T. Nakamura and H. Tsukagoshi, "Proposal of Soft Slip-in Manipulator Capable of Sliding Under The Human Body," 10th JFPS International Symposium on Fluid Power, p. 6, 2017.

[12] J. D. Greer, T. K. Morimoto, A. M. Okamura, and E. W. Hawkes, "A Soft, Steerable Continuum Robot That Grows via Tip Extension," Soft Robotics, vol. 6, no. 1, pp. 95–108, Feb. 2019, doi: 10.1089/soro.2018.0034.

[13] K. Hosaka, H. Tsukagoshi, and A. Kitagawa, "MOBILE ROBOT BY A DRAWING-OUT TYPE ACTUATOR FOR SMOOTH LOCOMOTION INSHIDE NARROW AND CURVING PIPES," Proceedings of the 8th JFPS International Symposium on Fluid Power, OKINAWA 2011Oct. 25-28, 2011 p. 6, 2011.

[14] E. W. Hawkes, L. H. Blumenschein, J. D. Greer, and A. M. Okamura, "A soft robot that navigates its environment through growth," Sci. Robot., vol. 2, no. 8, p. eaan3028, Jul. 2017, doi: 10.1126/scirobotics.aan3028.

[15] J. D. Greer, T. K. Morimoto, A. M. Okamura, and E. W. Hawkes, "A Soft, Steerable Continuum Robot That Grows via Tip Extension," Soft Robotics, vol. 6, no. 1, pp. 95–108, Feb. 2019, doi: 10.1089/soro.2018.0034.

[16] L. H. Blumenschein, L. T. Gan, J. A. Fan, A. M. Okamura, and E. W. Hawkes, "A Tip-Extending Soft Robot Enables Reconfigurable and Deployable Antennas," IEEE Robot. Autom. Lett., vol. 3, no. 2, pp. 949–956, Apr. 2018, doi: 10.1109/LRA.2018.2793303.

[17] C. Lucarotti, M. Totaro, A. Sadeghi, B. Mazzolai, and L. Beccai, "Revealing bending and force in a soft body through a plant root inspired approach," Sci Rep, vol. 5, no. 1, p. 8788, Aug. 2015, doi: 10.1038/srep08788.

[18] K. Tadakuma et al., "Nemertea Proboscis Inspired Extendable Mechanism," 30th 2019 International Symposium on Micro-Nano Mechatronics and Human Science, Dec.1-4, 2019.

[19] R. Gibson, "A NEW GENUS AND SPECIES OF LINEID HETERONEMERTEAN FROM SOUTH AFRICA, POLYBRACHIORHYNCHUS DAYI (NEMERTEA: ANOPLA), POSSESSING A MULTI- BRANCHED PROBOSCIS," BULLETIN OF MARINE SCIENCE, vol. 27, p. 20, 1977.

[20] H. Kajihara, "A histology-free description of the branched-proboscis ribbonworm Gorgonorhynchus albocinctus sp. nov. (Nemertea: Heteronemertea)," Publ. Seto Mar. Biol. Lab., 43: 92–102, 2015

[21] Animal Wire/"ALIEN WORM SHOOTS GOOEY WEB": In Initial Caps. (May, 15, 2015). Accessed:Feb. 22, 2020. [Online Video]. Available:https://www.youtube.com/watch?v=MmGz8gotCPs

[22] "AVID TV. Nemertea marine ribbon worm / น่ากลัว หนอนทะเล เป็นแบบนี้ In Initial Caps. (June. 3. 2015). Accessed : November. 8. 2019 [Online Video]. Available: https://www.youtube.com/watch?v=ez4l514lBb8"

[23] J. Luong et al., "Eversion and Retraction of a Soft Robot Towards the Exploration of Coral Reefs," in 2019 2nd IEEE International Conference on Soft Robotics (RoboSoft), Seoul, Korea (South), 2019, pp. 801–807, doi: 10.1109/ROBOSOFT.2019.8722730.

[24] M. M. Coad, R. P. Thomasson, L. H. Blumenschein, N. S. Usevitch, E. W. Hawkes, and A. M. Okamura, "Retraction of Soft Growing Robots without Buckling," IEEE Robot. Autom. Lett., pp. 1–1, 2020, doi: 10.1109/LRA.2020.2970629.

[25] S.-G. Jeong et al., "A Tip Mount for Carrying Payloads using Soft Growing Robots," arXiv:1912.08297 [cs], Dec. 2019.